\documentstyle[aclap]{article}

\title{\vspace{-0.5in}Combining Unsupervised 
Lexical Knowledge Methods for 
Word Sense Disambiguation
\thanks{\,\,This research has been partially funded by CICYT TIC96-1243-C03-02
(ITEM project) and the European Comission LE-4003 (EuroWordNet project).}}
\author{German Rigau, Jordi Atserias \\
Dept. de Llenguatges i Sist. Inform\`{a}tics \\
Universitat Polit\`{e}cnica de Catalunya \\
Barcelona, Catalonia \\
{\tt \{g.rigau,batalla\}@lsi.upc.es}\And
Eneko Agirre \\
Lengoaia eta Sist. Informatikoak saila \\
Euskal Herriko Unibertsitatea \\
Donostia, Basque Country \\
{\tt jibagbee@si.ehu.es}}

\begin{document}
\bibliographystyle{fullname}
\maketitle
\vspace{-0.5in}
\begin{abstract}
This paper presents a method to combine a set of unsupervised algorithms
that can accurately disambiguate word senses in a large, completely untagged
corpus. Although most of the techniques for word sense resolution have been
presented as stand-alone, it is our belief that full-fledged lexical
ambiguity resolution should combine several information sources and
techniques. The set of techniques have been applied in a combined way to
disambiguate the genus terms of two machine-readable dictionaries (MRD),
enabling us to construct complete taxonomies for Spanish and French. Tested
accuracy is above 80\% overall and 95\% for two-way ambiguous genus terms,
showing that taxonomy building is not limited to structured dictionaries
such as LDOCE.
\end{abstract}

\section{Introduction}

While in English the ``lexical bottleneck'' problem \cite{Briscoe} seems to be
softened (e.g. WordNet \cite{Miller}, Alvey Lexicon \cite{Grover+al}, COMLEX
\cite{Grishman+al}, etc.) there are no available wide range lexicons for
natural language processing (NLP) for other languages. Manual construction
of lexicons is the most reliable technique for obtaining structured lexicons
but is costly and highly time-consuming. This is the reason for many
researchers having focused on the massive acquisition of lexical knowledge
and semantic information from pre-existing structured lexical resources as
automatically as possible.

As dictionaries are special texts whose subject matter is a language (or a
pair of languages in the case of bilingual dictionaries) they provide a wide
range of information about words by giving definitions of senses of words,
and, doing that, supplying knowledge not just about language, but about the
world itself.

One of the most important relation to be extracted from machine-readable
dictionaries (MRD) is the hyponym/hypernym relation among dictionary senses
(e.g. \cite{Amsler}, \cite{Vossen+Serail}
) not only because of its own importance as the backbone of taxonomies, but
also because this relation acts as the support of main inheritance
mechanisms helping, thus, the acquisition of other relations and semantic
features \cite{Cohen+Loiselle}, providing formal structure and avoiding
redundancy in the lexicon \cite{Briscoe+al}. For instance, following the
natural chain of dictionary senses described in the {\em Diccionario General
Ilustrado de la Lengua Espa\~{n}ola} \cite{DGILE} we can discover that a
{\em bonsai} is a cultivated plant or bush.

\begin{description} 
	\item[bonsai\_1\_2] \em planta y arbusto as\'{\i} cultivado. \\
		\rm (bonsai, plant and bush cultivated in that way)
\end{description}

The hyponym/hypernym relation appears between the entry word (e.g. {\em
bonsai}) and the genus term, or the core of the phrase (e.g. {\em planta}
and {\em arbusto}). Thus, usually a dictionary definition is written to
employ a genus term combined with differentia 
which distinguishes the word being defined from other words with the same
genus term\footnote{ For other kind of definition patterns not based on
genus, a genus-like term was added after studying those patterns.}.

As lexical ambiguity pervades language in texts, the words used in
dictionary are themselves lexically ambiguous. Thus, when constructing
complete disambiguated taxonomies, the correct dictionary sense of the genus
term must be selected in each dictionary definition, performing what is
usually called Word Sense Disambiguation (WSD)\footnote{ Called also Lexical
Ambiguity Resolution,
Word Sense Discrimination,
Word Sense Selection
or Word Sense Identification.
}.  In the previous example {\em planta} has thirteen senses and {\em
arbusto} only one. 

Although a large set of dictionaries have been exploited as lexical
resources, the most widely used monolingual MRD for NLP is LDOCE which was
designed for learners of English. It is clear that different dictionaries do
not contain the same explicit information. The information placed in LDOCE
has allowed to extract other implicit information easily, e.g. taxonomies
\cite{Bruce+al}. Does it mean that only highly structured dictionaries
like LDOCE are suitable to be exploited to provide lexical resources for NLP
systems?

We explored this question probing two disparate dictionaries: {\em
Diccionario General Ilustrado de la Lengua Espa\~{n}ola} \cite{DGILE} for
Spanish, and {\em Le Plus Petit Larousse} \cite{LPPL} for French. Both are
substantially poorer in coded information than LDOCE \cite{LDOCE}\footnote{ In LDOCE,
dictionary senses are explicitly ordered by frequency, 86\% dictionary
senses have semantic codes and 44\% of dictionary senses have pragmatic
codes.}.  These dictionaries are very different in number of headwords,
polysemy degree, size and length of definitions (c.f. table 1). While DGILE
is a good example of a large sized dictionary, LPPL shows to what extent the
smallest dictionary is useful.

\begin{table} \footnotesize \centering
\begin{tabular}{|l|r|r|r|r|} \hline
& \multicolumn{2}{c|}{DGILE} & \multicolumn{2}{c|}{LPPL} \\ \cline{2-5}
& overall 	& nouns		& overall	& nouns \\  \hline

headwords	
	& 93,484	& 53,799	& 15,953	& 10,506 \\ \hline
senses		
	& 168,779	& 93,275	& 22,899	& 13,740 \\ \hline
total number &&&& \\ of words	
	& 1,227,380	& 903,163	& 97,778	& 66,323 \\ \hline
average length &&&& \\ of definition	
	& 7.26		& 9.68		& 3.27		& 3.82 \\ \hline
\end{tabular}
\caption{Dictionary Data}
\end{table}

Even if most of the techniques for WSD are presented as stand-alone, it is
our belief, following the ideas of \cite{McRoy}, that full-fledged lexical
ambiguity resolution should combine several information sources and
techniques. This work does not address all the heuristics cited in her
paper, but profits from techniques that were at hand, without any claim of
them being complete. In fact we use unsupervised techniques, i.e. those that
do not require hand-coding of any kind, that draw knowledge from a variety
of sources -- the source dictionaries, bilingual dictionaries and WordNet -- in
diverse ways. This paper tries to proof that using an appropriate method to
combine those heuristics we can disambiguate the genus terms with reasonable
precision, and thus construct complete taxonomies from any conventional
dictionary in any language.

This paper is organized as follows. After this short introduction, section 2
shows the methods we have applied. Section 3 describes the test sets and
shows the results. Section 4 explains the construction of the lexical
knowledge resources used. Section 5 discusses previous work, and finally,
section 6 faces some conclusions and comments on future work.

\section{Heuristics for Genus Sense Disambiguation}

As the methods described in this paper have been developed for being applied
in a combined way, each one must be seen as a container of some part of the
knowledge (or heuristic) needed to disambiguate the correct hypernym
sense. Not all the heuristics are suitable to be applied to all
definitions. For combining the heuristics, each heuristic assigns each
candidate hypernym sense a normalized weight, i.e. a real number ranging
from 0 to 1 (after a scaling process, where maximum score is assigned 1,
c.f. section 2.9). The heuristics applied range from the simplest
(e.g. heuristic 1, 2, 3 and 4) to the most informed ones (e.g. heuristics 5,
6, 7 and 8), and use information present in the entries under study
(e.g. heuristics 1, 2, 3 and 4) or extracted from the whole dictionary as a
unique lexical knowledge resource (e.g. heuristics 5 and 6) or combining
lexical knowledge from several heterogeneous lexical resources
(e.g. heuristic 7 and 8).

\subsection{Heuristic 1: Monosemous Genus Term}

This heuristic is applied when the genus term is monosemous. As there is
only one hypernym sense candidate, the hyponym sense is attached to it. Only
12\% of noun dictionary senses have monosemous genus terms in DGILE, whereas
the smaller LPPL reaches 40\%.

\subsection{Heuristic 2: Entry Sense Ordering}

This heuristic assumes that senses are ordered in an entry by frequency of
usage. That is, the most used and important senses are placed in the entry
before less frequent or less important ones. This heuristic provides the
maximum score to the first sense of the hypernym candidates and decreasing
scores to the others.

\subsection{Heuristic 3: Explicit Semantic Domain}

This heuristic assigns the maximum score to the hypernym sense which has the
same semantic domain tag as the hyponym. This heuristic is of limited
application: LPPL lacks semantic tags, and less than 10\% of the definitions
in DGILE are marked with one of the 96 different semantic domain tags
(e.g. {\em med.} for medicine, or {\em der.} for law, etc.).

\subsection{Heuristic 4: Word Matching}

This heuristic trusts that related concepts will be expressed using the same
content words. Given two definitions -- that of the hyponym and that of one
candidate hypernym -- this heuristic computes the total amount of content
words shared (including headwords). Due to the morphological productivity of
Spanish and French, we have considered different variants of this
heuristic. For LPPL the match among lemmas proved most useful, while DGILE
yielded better results when matching the first four characters of words.

\subsection{Heuristic 5: Simple Cooccurrence}

This heuristic uses cooccurrence data collected from the whole dictionary
(see section 4.1 for more details). Thus, given a hyponym definition ($O$)
and a set of candidate hypernym definitions, this method selects the
candidate hypernym definition ($E$) which returns the maximum score given by
formula (1):

\begin{equation}	
SC(O,E) = \sum_{w_{i} \in O \wedge w_{j} \in E} cw(w_{i},w_{j})
\end{equation}

The cooccurrence weight ($cw$) between two words can be given by Cooccurrence
Frequency, Mutual Information \cite{Church+Hanks} or Association Ratio
\cite{Resnik:92}. We tested them using different context window sizes. Best
results were obtained in both dictionaries using the Association Ratio. In
DGILE window size 7 proved the most suitable, whereas in LPPL whole
definitions were used.

\subsection{Heuristic 6: Cooccurrence Vectors}

This heuristic is based on the method presented in \cite{Wilks+al} which
also uses cooccurrence data collected from the whole dictionary
(c.f. section 4.1). Given a hyponym definition ($O$) and a set of candidate
hypernym definitions, this method selects the candidate hypernym ($E$) which
returns the maximum score following formula (2):

\begin{equation}
CV(O,E) = sim(V_{O},V_{E})
\end{equation}

The similarity ($sim$) between two definitions can be measured by the dot
product, the cosine function or the Euclidean distance between two vectors
($V_{O}$ and $V_{E}$) which represent the contexts of the words presented in the
respective definitions following formula (3):

\begin{equation}
V_{Def} = \sum_{w_{i} \in Def} civ(w_{i})
\end{equation}

The vector for a definition ($V_{Def}$) is computed adding the cooccurrence
information vectors of the words in the definition ($civ(w_{i})$). The
cooccurrence information vector for a word is collected from the whole
dictionary using Cooccurrence Frequency, Mutual Information or Association
Ratio. The best combination for each dictionary vary: whereas the dot
product, Association Ratio, and window size 7 proved best for DGILE, the
cosine, Mutual Information and whole definitions were preferred for LPPL.

\subsection{Heuristic 7: Semantic Vectors}

Because both LPPL and DGILE are poorly semantically coded we decided to
enrich the dictionary assigning automatically a semantic tag to each
dictionary sense (see section 4.2 for more details). Instead of assigning
only one tag we can attach to each dictionary sense a vector with weights
for each of the 25 semantic tags we considered (which correspond to the 25
lexicographer files of WordNet \cite{Miller}). In this case,
given an hyponym ($O$) and a set of possible hypernyms we select the candidate
hypernym ($E$) which yields maximum similarity among semantic vectors:

\begin{equation}
SV(O,E) = sim(V_{O},V_{E})
\end{equation}

where $sim$ can be the dot product, cosine or Euclidean Distance, as
before. Each dictionary sense has been semantically tagged with a vector of
semantic weights following formula (5).

\begin{equation}
V_{Def} = \sum_{w_{i} \in Def} swv(w_{i})
\end{equation}

The salient word vector ($swv$) for a word contains a saliency weight
\cite{Yarowsky:92} for each of the 25 semantic tags of WordNet. Again, the
best method differs from one dictionary to the other: each one prefers the
method used in the previous section.

\subsection{Heuristic 8: Conceptual Distance}

Conceptual distance provides a basis for determining closeness in meaning
among words, taking as reference a structured hierarchical net. Conceptual
distance between two concepts is essentially the length of the shortest path
that connects the concepts in the hierarchy. In order to apply conceptual
distance, WordNet was chosen as the hierarchical knowledge base, and
bilingual dictionaries were used to link Spanish and French words to the
English concepts.

Given a hyponym definition ($O$) and a set of candidate hypernym definitions,
this heuristic chooses the hypernym definition ($E$) which is closest
according to the following formula:

\begin{equation}
CD(O,E) = dist(headword_{O},genus_{E})
\end{equation}

That is, Conceptual Distance is measured between the headword of the hyponym
definition and the genus of the candidate hypernym definitions using formula
(7), c.f. \cite{Agirre+al}. To compute the distance between any two words
($w_{1}$,$w_{2}$), all the corresponding concepts in WordNet ($c_{1_{i}}$,
$c_{2_{j}}$) are searched via a bilingual dictionary, and the minimum of the
summatory for each concept in the path between each possible combination of
$c_{1_{i}}$and $c_{2_{j}}$ is returned, as shown below:

\begin{equation}
\renewcommand\arraystretch{0.7}
dist(w_{1},w_{2}) = \min_{\begin{array}{c}  
                             \scriptstyle c_{1_{i}} \in w_{1} \\
                             \scriptstyle c_{2_{j}} \in w_{2}
			  \end{array}}
		        \sum_{\begin{array}{c}  
                             		\scriptstyle c_{k} \in \\
					\scriptstyle path(c_{1_{i}},
							      c_{2_{j}})
			       \end{array}}
                            \frac{1}{depth(c_{k})} 
\end{equation}

Formulas (6) and (7) proved the most suitable of several other possibilities
for this task, including those which included full definitions in (6) or
those using other Conceptual Distance formulas, c.f. \cite{Agirre+Rigau}.

\subsection{Combining the heuristics: Summing}

As outlined in the beginning of this section, the way to combine all the
heuristics in one single decision is simple. The weights each heuristic
assigns to the rivaling senses of one genus are normalized to the interval
between 1 (best weight) and 0. Formula (8) shows the normalized value a
given heuristic will give to sense $E$ of the genus, according to the weight
assigned to the heuristic to sense $E$ and the maximum weight of all the
sense of the genus $E_{i}$.

\begin{equation}
vote(O,E) = \frac{weight(O,E)}{\max_{E_{i}} \left(weigth(O,E_{i})\right)}
\end{equation}

The values thus collected from each heuristic, are added up for each
competing sense. The order in which the heuristics are applied has no
relevance at all.

\section{Evaluation}

\subsection{Test Set}

In order to test the performance of each heuristic and their combination, we
selected two test sets at random (one per dictionary): 391 noun senses for
DGILE and 115 noun senses for LPPL, which give confidence rates of 95\% and
91\% respectively. From these samples, we retained only those for which the
automatic selection process selected the correct genus (more than 97\% in
both dictionaries). Both test sets were disambiguated by hand. Where
necessary multiple correct senses were allowed in both dictionaries. Table 2
shows the data for the test sets.

\begin{table} \footnotesize \centering
\begin{tabular}{|l|r|r|} \hline
& \multicolumn{1}{c|}{DGILE} & \multicolumn{1}{c|}{LPPL} \\ \hline
Test Sampling 			& 391 		& 115	\\ \hline
Correct Genus Selected 		& 382 (98\%) 	& 111 (97\%)	\\ \hline
Monosemous 			& 61 (16\%) 	& 40 (36\%)	\\ \hline
Senses per genus	 	& 2.75 	& 2.29	        \\ \hline
idem (polysemous only) 		& 3.64 	& 3.02 		\\ \hline
Correct senses per genus	& 1.38 	& 1.05	        \\ \hline
idem (polysemous only) 		& 1.51 	& 1.06 		\\ \hline
\end{tabular}
\caption{Test Sets}
\end{table}

\subsection{Results}

Table 3 summarizes the results for polysemous genus.

\begin{table*} \footnotesize \centering
\begin{tabular}{|l|r|r|r|r|r|r|r|r|r|r|} \hline
LPPL		& random & (1)	& (2)	& (3)	& (4)	& (5)	& (6)	& (7)	& (8)	& Sum \\ \hline
recall 		& 36\%	& -	& 66\%	& -	& 8\%	& 11\%	& 22\%	& 11\%	& 50\%	& 73\% \\ \hline
precision 	& 36\%	& -	& 66\%	& -	& 66\%	& 44\%	& 61\%	& 57\%	& 76\%	& 73\% \\ \hline
coverage 	& 100\%	& -	& 100\%	& -	& 12\%	& 25\%	& 36\%	& 19\%	& 66\%	& 100\% \\ \hline
\multicolumn{11}{|l|}{DGILE}  \\ \hline
recall 		& 30\%	& -	& 70\%	& 1\%	& 44\%	& 57\%	& 60\%	& 57\%	& 47\%	& 79\% \\ \hline
precision 	& 30\%	& -	& 70\%	& 100\%	& 72\%	& 57\%	& 60\%	& 58\%	& 49\%	& 79\% \\ \hline
coverage 	& 100\%	& -	& 100\%	& 1\%	& 61\%	& 100\%	& 100\%	& 99\%	& 95\%	& 100\% \\ \hline
\end{tabular}
\caption{Results for polysemous genus.}
\end{table*}

In general, the results obtained for each heuristic seem to be poor, but
always over the random choice baseline (also shown in tables 3 and 4). The
best heuristics according to the recall in both dictionaries is the sense
ordering heuristic (2). For the rest, the difference in size of the
dictionaries could explain the reason why cooccurrence-based heuristics (5
and 6) are the best for DGILE, and the worst for LPPL. Semantic distance
gives the best precision for LPPL, but chooses an average of 1.25 senses for
each genus. 

With the combination of the heuristics (Sum) we obtained an improvement over
sense ordering (heuristic 2) of 9\% (from 70\% to 79\%) in DGILE, and of
7\% (from 66\% to 73\%) in LPPL, maintaining in both cases a coverage of
100\%. Including monosemous genus in the results (c.f. table 4), the sum is
able to correctly disambiguate 83\% of the genus in DGILE (8\% improvement
over sense ordering) and 82\% of the genus in LPPL (4\% improvement). Note
that we are adding the results of eight different heuristics with eight
different performances, improving the individual performance of each one.

\begin{table*} \footnotesize \centering
\begin{tabular}{|l|r|r|r|r|r|r|r|r|r|r|} \hline
LPPL 	& random	& (1)	& (2)	& (3)	& (4)	& (5)	& (6)	& (7)	& (8)	& Sum	 \\ \hline
recall		& 59\%	& 35\%	& 78\%	& -	& 40\%  & 42\%	& 50\%	& 42\%	& 68\%	& 82\% \\ \hline
precision	& 59\%	& 100\%	& 78\%	& -	& 93\%  & 82\%	& 84\%	& 88\%	& 87\%	& 82\% \\ \hline
coverage	& 100\%	& 35\%	& 100\%	& -	& 43\%  & 51\%	& 59\%	& 48\%	& 78\%	& 100\% \\ \hline
\multicolumn{11}{|l|}{DGILE}  \\ \hline
recall 		& 41\%	& 16\%	& 75\%	& 2\%	& 41\%	& 59\%	& 63\%	& 59\%	& 48\%	& 83\% \\ \hline
precision	& 41\%	& 100\%	& 75\%	& 100\%	& 79\%	& 65\%	& 66\%	& 63\%	& 57\%	& 83\% \\ \hline
coverage	& 100\%	& 16\%	& 100\%	& 2\%	& 56\%	& 95\%	& 97\%	& 94\%	& 89\%	& 100\% \\ \hline
\end{tabular}
\caption{Overall results.}
\end{table*}

In order to test the contribution of each heuristic to the total knowledge,
we tested the sum of all the heuristics, eliminating one of them in
turn. The results are provided in table 5.

\begin{table*}[!htb] \footnotesize \centering
\begin{tabular}{|l|r|r|r|r|r|r|r|r|r|r|} \hline
LPPL		& Sum	& -(1)	& -(2)	& -(3)	& -(4)	& -(5)	& -(6)	& -(7)	& -(8)  \\ \hline
recall 		& 82\%	& 73\%	& 74\%	& -	& 73\%	& 76\%	& 77\%	& 77\%	& 78\%  \\ \hline
precision	& 82\%	& 73\%	& 75\%	& -	& 73\%	& 76\%	& 77\%	& 77\%	& 78\%  \\ \hline
coverage	& 100\%	& 100\%	& 99\%	& -	& 100\%	& 100\%	& 100\%	& 100\%	& 100\%  \\ \hline
\multicolumn{10}{|l|}{DGILE}  \\ \hline
recall  	& 83\% 	& 79\% 	& 72\% 	& 81\% 	& 81\% 	& 81\% 	& 81\% 	& 81\% 	& 77\% \\ \hline
precision 	& 83\% 	& 79\% 	& 72\% 	& 82\% 	& 81\% 	& 81\% 	& 81\% 	& 81\% 	& 77\% \\ \hline
coverage 	& 100\% & 100\% & 100\% & 98\% 	& 100\% & 100\% & 100\% & 100\% & 100\% \\ \hline
\end{tabular}
\caption{Knowledge provided by each heuristic (overall results).}
\end{table*}

\cite{Gale+al:93} estimate that any sense-identification system that does not
give the correct sense of polysemous words more than 75\% of the time would
not be worth serious consideration. As table 5 shows this is not the case in
our system. For instance, in DGILE heuristic 8 has the worst performance
(see table 4, precision 57\%), but it has the second larger contribution
(see table 5, precision decreases from 83\% to 77\%). That is, even those
heuristics with poor performance can contribute with knowledge that other
heuristics do not provide.

\subsection{Evaluation}

The difference in performance between the two dictionaries show that quality
and size of resources is a key issue. Apparently the task of disambiguating
LPPL seems easier: less polysemy, more monosemous genus and high precision
of the sense ordering heuristic. However, the heuristics that depend only on
the size of the data (5, 6) perform poorly on LPPL, while they are powerful
methods for DGILE.

The results show that the combination of heuristics is useful, even if the
performance of some of the heuristics is low. The combination performs
better than isolated heuristics, and allows to disambiguate all the genus of
the test set with a success rate of 83\% in DGILE and 82\% in LPPL.

All the heuristics except heuristic 3 can readily be applied to any other
dictionary. Minimal parameter adjustment (window size, cooccurrence weigth
formula and vector similarity function) should be done to fit the
characteristics of the dictionary, but according to our results it does not
alter significantly the results after combining the heuristics.

\section{Derived Lexical Knowledge Resources}

\subsection{Cooccurrence Data}

Following \cite{Wilks+al} two words cooccur if they appear in the same
definition (word order in definitions are not taken into account). For
instance, for DGILE, a lexicon of 300,062 cooccurrence pairs among 40,193
word forms was derived (stop words were not taken into account). Table 6
shows the first eleven words out of the 360 which cooccur with {\em vino} (wine)
ordered by Association Ratio. From left to right, Association Ratio and
number of occurrences.

\begin{table} \footnotesize \centering
\begin{tabular}{|r|r|l|} \hline
AR 	& \#oc. 	& \\ \hline
11.1655	& 15 	& {\em tinto} (red) \\ \hline
10.0162	& 23	& {\em beber} (to drink) \\ \hline
9.6627	& 14	& {\em mosto} (must) \\ \hline
8.6633	& 9	& {\em jerez} (sherry) \\ \hline
8.1051	& 9	& {\em cubas} (cask, barrel) \\ \hline
8.0551	& 16	& {\em licor} (liquor) \\ \hline
7.2127	& 17	& {\em bebida} (drink) \\ \hline
6.9338	& 12	& {\em uva} (grape) \\ \hline
6.8436	& 9	& {\em trago} (drink, swig) \\ \hline
6.6221	& 12	& {\em sabor} (taste) \\ \hline
6.4506	& 15	& {\em pan} (bread) \\ \hline
\end{tabular}
\caption{Example of association ratio for {\em vino} (wine).}
\end{table}

The lexicon (or machine-tractable dictionary, MTD) thus produced from the
dictionary is used by heuristics 5 and 6.

\subsection{Multilingual Data}

Heuristics 7 and 8 need external knowledge, not present in the dictionaries
themselves. This knowledge is composed of semantic field tags and
hierarchical structures, and both were extracted from WordNet. In order to
do this, the gap between our working languages and English was filled with
two bilingual dictionaries. For this purpose, we derived a list of links for
each word in Spanish and French as follows.

Firstly, each Spanish or French word was looked up in the bilingual
dictionary, and its English translation was found. For each translation
WordNet yielded its senses, in the form of WordNet concepts (synsets). The
pair made of the original word and each of the concepts linked to it, was
included in a file, thus producing a MTD with links between Spanish or
French words and WordNet concepts. Obviously some of this links are not
correct, as the translation in the bilingual dictionary may not necessarily
be understood in its senses (as listed in WordNet). The heuristics using
these MTDs are aware of this.

For instance when accessing the semantic fields for {\em vin} (French) we
get a unique translation, wine, which has two senses in WordNet:
\verb|<wine,vino>| as a beverage, and \verb|<wine, wine-coloured>| as a kind
of color. In this example two links would be produced \verb|(|{\em vin}\verb|,
<wine,vino>)| and \verb|(|{\em vin}\verb|, <wine, wine-coloured>)|. This link
allows us to get two possible semantic fields for {\em vin}
(\verb|noun.food|, file 13, and \verb|noun.attribute|, file 7) and the whole
structure of the hierarchy in WordNet for each of the concepts.

\section{Comparison with Previous Work}

Several approaches have been proposed for attaching the correct sense (from
a set of prescribed ones) of a word in context. Some of them 
have been fully tested in real size texts (e.g. statistical methods
\cite{Yarowsky:92}, \cite{Yarowsky:94}, \cite{Miller+Teibel}, knowledge
based methods \cite{Sussna}, \cite{Agirre+Rigau}, or mixed methods
\cite{Richardson+al}, \cite{Resnik:95}). The performance of WSD is reaching
a high stance, although usually only small sets of words with clear sense
distinctions are selected for disambiguation (e.g. \cite{Yarowsky:95}
reports a success rate of 96\% disambiguating twelve words with two clear
sense distinctions each one).

This paper has presented a general technique for WSD which is a combination
of statistical and knowledge based methods, and which has been applied to
disambiguate all the genus terms in two dictionaries.

Although this latter task could be seen easier than general WSD\footnote{In
contrast to other sense distinctions Dictionary word senses frequently
differ in subtle distinctions (only some of which have to do with meaning
\cite{Gale+al:93}) producing a large set of closely related dictionary
senses \cite{Jacobs}.}, genus are usually frequent and general words with
high ambiguity\footnote{ However, in dictionary definitions the headword and
the genus term have to be the same part of speech.}.  While the average of
senses per noun in DGILE is 1.8 the average of senses per noun genus is 2.75
(1.30 and 2.29 respectively for LPPL). Furthermore, it is not possible to
apply the powerful ``one sense per discourse'' property \cite{Yarowsky:95}
because there is no discourse in dictionaries.

WSD is a very difficult task even for humans\footnote{ 
\cite{Wilks+al} disambiguating 197 occurrences of the word bank in LDOCE
say ``was not an easy task, as some of the usages of bank did not seem to
fit any of the definitions very well''. Also \cite{Miller+al} tagging
semantically SemCor by hand, measure an error rate around 10\% for
polysemous words.}, 
but semiautomatic techniques to disambiguate genus have
been broadly used \cite{Amsler} \cite{Vossen+Serail} \cite{Ageno+al}
\cite{Artola} and some attempts to do automatic genus
disambiguation have been performed using the semantic codes of the
dictionary 
\cite{Bruce+al} or using cooccurrence data
extracted from the dictionary itself \cite{Wilks+al}.

Selecting the correct sense for LDOCE genus terms, \cite{Bruce+al}) report
a success rate of 80\% (90\% after hand coding of ten genus). This impressive
rate is achieved using the intrinsic characteristics of LDOCE. Furthermore,
using only the implicit information contained into the dictionary
definitions of LDOCE \cite{Cowie+al} report a success rate of 47\% at a
sense level. \cite{Wilks+al} reports a success rate of 45\% disambiguating
the word bank (thirteen senses LDOCE) using a technique similar to
heuristic 6. In our case, combining informed heuristics and without explicit
semantic tags, the success rates are 83\% and 82\% overall, and 95\% and 75\%
for two-way ambiguous genus (DGILE and LPPL data, respectively). Moreover, 93\%
and 92\% of times the real solution is between the first and second proposed
solution.

\section{Conclusion and Future Work}

The results show that computer aided construction of taxonomies using
lexical resources is not limited to highly-structured dictionaries as LDOCE,
but has been succesfully achieved with two very different dictionaries. All
the heuristics used are unsupervised, in the sense that they do not need
hand-codding of any kind, and the proposed method can be adapted to any
dictionary with minimal parameter setting.

Nevertheless, quality and size of the lexical knowledge
resources are important. As the results for LPPL show, small dictionaries
with short definitions can not profit from raw corpus techniques (heuristics
5, 6), and consequently the improvement of precision over the random
baseline or first-sense heuristic is lower than in DGILE. 

We have also shown that such a simple technique as just summing is a
useful way to combine knowledge from several unsupervised WSD methods,
allowing to raise the performance of each one in isolation (coverage and/or
precision). Furthermore, even those heuristics with apparently poor results
provide knowledge to the final result not provided by the rest of
heuristics. Thus, adding new heuristics with different methodologies and
different knowledge (e.g. from corpora) as they become available will
certainly improve the results. 

Needless to say, several improvements can be done both in individual
heuristic and also in the method to combine them. For instance, the
cooccurrence heuristics have been applied quite indiscriminately, even in
low frequency conditions. Significance tests or association coefficients
could be used in order to discard low confidence decisions. Also, instead of
just summing, more clever combinations can be tried, such as training
classifiers which use the heuristics as predictor variables. 

Although we used these techniques for genus disambiguation we
expect similar results (or even better taken the ``one sense per discourse''
property and lexical knowledge acquired from corpora) for
the WSD problem.

\section{Acknowledgments}
This work would not be possible without the collaboration of our colleagues,
specially Jose Mari Arriola, Xabier Artola, Arantza Diaz de Ilarraza, Kepa
Sarasola and Aitor Soroa in the Basque Country and Horacio Rodr\'{i}guez in
Catalonia.

\end{document}